\documentclass[11pt]{article}

\usepackage[utf8]{inputenc}
\usepackage[T1]{fontenc}
\usepackage[margin=1in]{geometry}
\usepackage{amsmath,amssymb,amsthm}
\usepackage{booktabs}
\usepackage{graphicx}
\usepackage{xcolor}
\usepackage{pifont}
\usepackage{microtype}
\usepackage{url}
\usepackage[numbers,sort&compress]{natbib}
\usepackage{hyperref}
\usepackage{cleveref}
\usepackage{authblk}
\usepackage{placeins}
\usepackage{float}
\usepackage{amsmath,amssymb}
\usepackage{needspace}

\hypersetup{colorlinks=true,linkcolor=blue!50!black,citecolor=blue!50!black,urlcolor=blue!50!black}

\theoremstyle{plain}
\newtheorem{theorem}{Theorem}

\theoremstyle{definition}

\newcommand{\Corr}{\mathrm{Corr}}

\newcommand{\cmark}{\textcolor{green!55!black}{\ding{51}}}
\newcommand{\xmark}{\textcolor{red!75!black}{\ding{55}}}

\newcommand{\Aone}{EIA}

\title{\textbf{Memory Reward Inflation in Self-Improving LLM Agents}}

\author[1]{Mohammad Asadolahi}
\author[2]{Amir Amini}
\author[3]{Samira Talebi}
\author[4]{Amirfarhad Farhadi}
\author[5]{Azadeh Zamanifar}
\affil[1]{\small Department of Computer Science and Engineering, University Of North Texas, TX, USA \\ \texttt{MohammadAsadolahi@my.unt.edu}}
\affil[2]{\small Department of Computer Science, Edge Hill University, UK \\ \texttt{AmirAmini@outlook.com}}
\affil[3]{\small Department of Computer Science, University of Cincinnati, OH, USA \\ \texttt{talebisa@mail.uc.edu}}
\affil[4]{\small School of Computer Engineering, Iran University of Science and Technology, Tehran, Iran \\ \texttt{am\_farhadi@mail.iust.ac.ir}}
\affil[5]{\small Department of Computer Engineering, SR.C, Islamic Azad University, Tehran, Iran \\ \texttt{azamanifar@iau.ac.ir}}
\date{June 2026}

\begin{document}
\maketitle

\begin{abstract}
Self-improving LLM agents increasingly learn from experience without updating any weights. Each episode is stored in an external memory, scored, and retrieved for similar future tasks to shape later behavior. Viewed through a reward lens, the stored score is a proxy reward for an implicit, non-parametric policy. Each retrieved episode then becomes a policy-improvement step whose reliability hinges on how that score is produced. In deployment, ground-truth labels are unavailable, so the stored reward is at best an LLM assessment. This substitution creates a failure mode, the \emph{Echo Gap}, across the memory baseed self-improving agents and model families studied. Incorrect episodes receive inflated rewards; thus, the agent preferentially reuses the very mistakes it has most confident in. Because the error compounds through memory rather than averaging out and the confirming judge's errors remain correlated with the original self-grading bias, so it cannot identify which memories are overvalued. The missing property is formalized as the \emph{Error-Independence Assumption} (EIA), which we prove is a \emph{necessary} condition for correcting the inflation, not merely a description of a good verifier: a usable signal must track truth \emph{and} decorrelate its error from the memory bias, and the recoverable payoff is a closed-form function of exactly those two quantities. We further show the inflation compounds not only when retrieval ranks by the stored score but also under plain similarity retrieval which is the regime the deployed agent uses. Finally, the answer-free de-inflation algorithm LUCID delivers a consistent end-to-end gain on the BIRD text-to-SQL benchmark. It raises execution accuracy to $56.9\%$, above both a Memento-style self-graded agent ($54.0\%$, a $+2.9$-point mean gain across seeds) and a memory-less agent of identical architecture ($52.4\%$).
\end{abstract}


\section{Introduction}
\label{sec:introduction}

Large Language Model (LLM) agents are increasingly designed to improve their behavior from past experience without updating any model parameter. Improvement comes not from fine-tuning but from a memory: the agent stores past episodes, assigns each a utility score, retrieves the episodes judged for similar future tasks, and writes each new episode back into the bank. Although the LLM stays parametrically fixed, the agent future behavior changes through the memories it chooses to use. This is attractive because it offers online adaptation without fine-tuning, personalization without retraining, and reusable experience without changing the underlying model. Viewed through a reward lens, such a memory-augmented agent is a memory decision process over stored experience in which retrieval plays the role of an implicit, non-parametric policy \citep{sutton2018reinforcement,reflexion2023,selfrefine2023,memento2025}. This lens is used here to study reward integrity rather than policy optimization.\footnote{Code, data, and per-episode memory traces are available at \url{https://github.com/MohammadAsadolahi/Reliable-Memory-Agents-in-the-Wild}.}

Because the stored score decides how memories are used, imitated or avoided, it acts as a reward signal, and label-free memory learning becomes a proxy-reward problem. The agent optimizes future behavior using a signal that may reward what merely \emph{appears} successful rather than what is genuinely correct. Its effectiveness therefore depends on whether the stored utility score reflects true task success.

In benchmark settings, an episode's score can often be derived from ground-truth labels: a factual answer is checked against a provided gold answer, a program against tests, a database query by execution. In real world problems, however, such labels are usually unavailable when the agent writes to memory. The practical substitute is LLM grading: the agent, or a closely related LLM judge, assigns the utility score and decides whether an episode is worth storing before its true success is known \citep{christiano2017deep,ouyang2022training,llmjudge2023}. The closest known relative is reward hacking, where a generator that also grades itself raises its self-assessed score without improving true quality \citep{rewardhacking2024}. The Echo Gap is the memory-specific and more dangerous form: the inflated score is not transient but written into persistent storage, and because these memories are retrieved, imitated, the error compounds through reuse rather than averaging out. This makes inflation a recurring risk for agents that evolve memory from experience without labels, across the models studied here. The agent does not merely store noisy memories; it retrieves and reinforces the ones it has most overvalued, so a label-free memory loop can severely magnify the model's own blind spots.

In the banks studied in this paper, more capable re-graders are more truth-aligned, yet their errors still correlate with the original self-grade bias: a verifier that shares the generator's blind spots can reproduce the same ranking distortion even when it is more capable overall.

This missing property is formalized as the \emph{Error-Independence Assumption} (\Aone{}). Let $V_i$ be a verifier score and $\nu_i=V_i-U_i$ the verifier error. In utility correction, where a verifier is incorporated into the stored memory score, a useful verifier must estimate correctness reliably and avoid errors that are correlated with the original memory bias. Formally, $\mathrm{Corr}(V,U)$ should be maximized, while $|\mathrm{Corr}(\nu,b)|$ should be minimized. This explains why de-correlated information channels, such as retrieval-based factual checking or execution-based tests, can provide useful corrective signals when parametric re-grading remains biased. It also clarifies the role of the negative result reported here: parametric judging is not claimed to be impossible in general; rather, in the label-free memory setting studied here, capability alone did not suffice to provide the error de-
correlation needed for reliable correction.

This mechanism is first studied at the level of the stored utilities themselves, showing that label-blind global calibration cannot correct heterogeneous, per-memory reward inflation because a global map of stored utilities cannot detect which memories are overvalued. Per-memory demotion is then shown to improve the bank when the verifier satisfies \Aone{}. This yields a principled correction rule for utility updates: do not merely rescale the memory scores; use a memory-specific signal whose errors do not echo the original bias.

The second part examines whether the Echo Gap translates into behavioral degradation in a full end-to-end memory agent. This part proceeds cautiously, with evidence on full BIRD text-to-SQL, a third-party benchmark with official execution-accuracy evaluation, using a faithful Memento-style retrieve--inject--write loop in which memories are written and reused for similar tasks without access to ground-truth labels. The proposed method, applies an answer-free de-inflation and consistently outperforms the self-graded baseline across seeds. 

The contributions are organized along three axes (phenomenon, diagnosis, and mitigation):

\begin{itemize}
\item \textbf{(C1) The Echo Gap.} We define and formalize the Echo Gap as frequency-correlated reward inflation in label-free self-improving LLM memory. It appears on live factual banks across model families, and, unlike a static grading error, the inflation compounds \emph{multiplicatively}, both through retrieval and, in the similarity-retrieval regime the deployed agent actually uses, through the trust the planner places in inflated memories (Theorem~\ref{thm:amplification}). 
\item \textbf{(C2) A diagnosis of why correction is hard.} A dynamical analysis of the write-back loop shows that self-grade inflation drives the bank to a \emph{corrupted attractor} matches the observed bank corruption  on the real agent. We prove the Error-Independence Assumption (\Aone{}) is a \emph{necessary} condition for de-inflation: the recoverable payoff is a closed-form function of exactly its two axes, so a verifier whose error echoes the self-grade cannot help at any step size, and stronger or different-family re-grading, judge ensembles, and generic memory filtering (confidence thresholding, self-consistency pruning, budget-matched random pruning) do not supply the required decorrelation on these banks.
\item \textbf{(C3) LUCID} The paper introduces LUCID (Leniency-corrected Utility Calibration via Independent Debiasing) algorithm, an answer-free general de-inflation on memory agents, and presents end-to-end evidence on BIRD text-to-SQL: without any ground-truth label in the loop, it improves a Memento-style memory agent over tested baselines, increasing accuracy due to reward de-inflation.
\end{itemize}

The broader implication is that label-free self-improvement should be understood as a reward-design problem where do do not have a gold label in the real world settings. Reliable self-improving agents therefore require reward evidence with failure modes that differ from those of the agent itself: de-correlated signals for memory reward correction.

% ============================================================================
\section{Related Work}
\label{sec:related}
% ============================================================================

\paragraph{Memory-augmented and self-improving agents.}
Prior work studies LLM agents that improve without updating model parameters by storing, retrieving, and reusing past experience, including episodic memory systems, case-based memory, and reflective agents that use feedbacks or self-refinement \citep{memento2025,reflexion2023,selfrefine2023}. These systems build on broader foundations for LLM agents and retrieval: interleaved reasoning-and-acting \citep{react2023}, chain-of-thought prompting \citep{cot2022}, retrieval-augmented generation \citep{rag2020}, and long-horizon memory for simulated agents \citep{genagents2023}. A closely related recent system, ReasoningBank, distills reasoning strategies from self-judged successful and failed episodes and reports gains from memory-driven self-evolution \citep{reasoningbank2025}. These results do not contradict such gains; they identify a regime boundary. Self-graded memory can help when ``looks right'' tracks ``is correct,'' and becomes risky when the two diverge. The present work focuses on a dependency that cuts across these designs: when stored episodes are scored by LLM-grades, that score becomes a reward-like signal that governs which episodes are later trusted and reused.

\paragraph{Reinforcement learning over agent memory.}
A growing line of work casts memory management itself as a learned policy, optimizing what to store, retrieve, or discard with reinforcement learning over outcome rewards \citep{memrl2026,memoryr1_2025,memalpha2025,agemem2026}, or framing retrieval as a contextual-bandit decision \citep{rsbandit2026}. These methods \emph{learn} a memory policy under an assumed-reliable reward. The present contribution is upstream and complementary: when that reward is a LLM-grade it is systematically inflated, and an answer-free signal repairs it. Reliable reward integrity is a prerequisite for any reward-driven memory learner, so the Echo Gap and its correction apply directly to these systems.

\paragraph{Self-evaluation and LLM-as-a-judge.}
Self-consistency, LLM-as-a-judge methods, self-refinement, and process-reward-style judging use model-generated judgments as label-free or weakly supervised signals \citep{selfconsistency2022,llmjudge2023,selfrefine2023,prmthink2025}. In many settings these judgments are used locally for selection, reranking, or critique. In persistent memory, however, the same judgment has a stronger role: it determines which past episodes are retrieved later. The relevant question is therefore not only whether a judge is accurate on average, but whether its errors are decorrelated from the original self-grade bias. Prior work documents that LLM judges exhibit position, verbosity, and self-preference biases \citep{wataoka2024,selfpref2025}, consistent with the finding here that stronger or different-family re-graders do not automatically repair self-graded memory. Relatedly, panels of such judges collapse to a fraction of their nominal independent votes when their errors are correlated \citep{ninejudges2026}, a panel-level form of the decorrelation \Aone{} requires.

\paragraph{Calibration, reward mis-specification, and verification.}
Post-hoc calibration can improve the numerical interpretation of scores, but it does not solve heterogeneous, per-memory self-grade inflation when the correction is label-blind. This connects our setting to reward mis-specification and reward hacking, where a proxy reward can be repeatedly optimized despite being misaligned with true utility \citep{rewardhacking2024}. A parallel literature shows that useful feedback can come from non-gold signals such as execution, tests, static analysis, metamorphic testing, retrieval evidence, and agreement checks \citep{quickcheck2000,csmith2011,metamorphic1998,metamorphicsurvey2016,codet2022}. The framework developed here explains when such signals are useful for memory correction: their failure modes must differ from the model's own self-grade.

\paragraph{Memory poisoning, repair, and intrinsic self-correction.}
Adversarial memory-poisoning work studies settings where external attackers inject misleading information into agent memory \citep{sedm2025,amemguard2025,memma2026}. The setting here is different: the corruption is endogenous. The agent writes its own wrong experience with an inflated score, and retrieval later amplifies that mistake. This is also related to evidence that LLM-maintained memories can degrade under continual updating \citep{faultymemory2026} and to work showing that LLMs often fail to self-correct without external feedback \citep{cannotselfcorrect2024}. More broadly, label-free evaluation is underidentified without assumptions on error independence or decorrelation \citep{jaffe2015,ntqr2023}\citep{bird2023}.; \Aone{} is a memory-specific instance of this requirement. Self-grade inflation is itself an instance of the broad reward-misalignment family, and is not claimed here as a new species of bias. The Echo Gap's contribution is mechanistic and is summarized in Table~\ref{tab:positioning}: among these failure modes it is the only one that is at once endogenous, label-free, compounds through retrieval and reuse rather than acting once (Theorem~\ref{thm:amplification}), and is paired with a correction criterion (\Aone{}) we prove necessary for de-inflation. The novelty is therefore not the existence of inflation but its loop-amplified, correctable form in self-improving memory agents.

\begin{table}[t]
\centering\small
\begin{tabular}{@{}lcccc@{}}
\toprule
Phenomenon & endogenous & label-free & compounds via reuse & correction criterion \\
\midrule
Reward hacking (RL) \citep{rewardhacking2024}           & \xmark & \xmark & opt.\ drift & --- \\
LLM-as-a-judge / self-preference \citep{wataoka2024,selfpref2025} & \xmark & \cmark & \xmark & --- \\
Self-correction failure \citep{cannotselfcorrect2024}   & \cmark & \cmark & \xmark & --- \\
Memory poisoning \citep{amemguard2025,memma2026}        & \xmark & --- & \cmark & defenses \\
Memory degradation \citep{faultymemory2026}             & \cmark & \cmark & partial & --- \\
\textbf{Echo Gap (ours)}                                & \cmark & \cmark & \cmark\,($e^{b/T}$) & \Aone{} \\
\bottomrule
\end{tabular}
\caption{The Echo Gap relative to known reward-misalignment phenomena. It is the only one that is simultaneously endogenous, arises in the label-free write regime, compounds multiplicatively through retrieval/reuse, via a ranking channel ($e^{b/T}$) and a trust channel that persists under similarity-only retrieval (Theorem~\ref{thm:amplification}), and comes with a criterion (\Aone{}), proved necessary for de-inflation, for which label-free signals correct it.}
\label{tab:positioning}
\end{table}

% ============================================================================
% ============================================================================
\section{Self-Graded Memory and the Echo Gap}
\label{sec:echo-gap}
% ============================================================================

This section studies label-free memory improvement in LLM agents through a persistent memory loop as past episodes are stored, scored, retrieved and exposed to the agent for new given tasks, and reused to influence future behavior. It defines the memory loop and its signals, formalizes the Echo Gap, and then shows that the failure mode appears in live factual memory banks. It also introduces the single correction principle used in the paper: \emph{answer-free de-inflation} for demoting the reward of memories that a de-correlated, oracle-free signal flags as inflated, so the agent stops trusting them as correct past experiences and may use them as wrong answer references to solve a given task.

\subsection{Memory Loop and Signals}
\label{subsec:memory-loop-signals}

A memory bank contains episodes
\begin{equation}\label{eq:memory}
m_i = (q_i, a_i, Q_i),
\end{equation}
where, in Equation~\eqref{eq:memory}, \(q_i\) is the task or query, \(a_i\) is the agent's response, and \(Q_i\) is the stored utility score attached to the episode. On a new task \(q\), the agent selects or exposes a set of prior episodes using a memory policy
\begin{equation}\label{eq:mempolicy}
\pi_{\mathrm{mem}}(m_i \mid q),
\end{equation}
The memory policy in Equation~\eqref{eq:mempolicy} may depend on semantic similarity, stored utility \(Q_i\), recency, task metadata, or a combination of these factors. The selected memories are injected into the context of agent, displayed as prior precedents, weighted by the planner, or otherwise made available to the agent before it produces the next response. The new episode is then scored and written back into the memory bank for future similar tasks.

The important point is that \(Q_i\) is not passive metadata. Once a stored score affects the agent's future responses, or even which memories are selected, trusted, or imitated, it becomes a reward-like signal for a non-parametric learning loop. The base model parameters remain frozen, but future behavior changes through the memories the agent chooses to reuse. In this sense, retrieval and reuse of past cases act as an implicit policy-improvement mechanism whose reliability depends on whether the stored utility score reflects true downstream value. To cover both cases, we use the general term \emph{reuse exposure} for the event that a memory is retrieved, injected, trusted and made influential for a later decision.

We focus on the label-free real world problems regime, in which the agent must write to memory when there is no ground-truth available. The assessment of whether a solution is right or wrong is instead produced by the agent itself, by a related model, or by an LLM-as-a-judge. This is the setting in which memory learning becomes vulnerable to proxy-reward failure: the system may reinforce a wrong solution that received a positive score, one that merely appears successful rather than being correct.

To analyze this loop, the deployment-time scores are distinguished from the post-hoc utility used only for measurement. Each memory \(m_i\) carries the three signals of Equation~\eqref{eq:signals}:
\begin{equation}\label{eq:signals}
r_i,\quad V_i,\quad U_i.
\end{equation}

The LLM grade \(r_i\) is the reward score assigned by an LLM at write time in label-free deployment, typically the agent grading its own episode (a \emph{self-grade}), or a closely related LLM judge. The verifier score \(V_i\) is an audit signal produced by some alternative channel, retrieved evidence, execution feedback, or invariant checks. The ground-truth utility \(U_i\) is used only for post-hoc measurement, oracle ceilings, and theorem statements. It is never used by the non-oracle memory loop.

Equation~\eqref{eq:bias} defines the self-grade bias \(b_i\) and the verifier error \(\nu_i\):
\begin{equation}\label{eq:bias}
b_i = r_i - U_i,
\qquad
\nu_i = V_i - U_i.
\end{equation}

The reuse count \(n_i\) denotes how often memory \(m_i\) is exposed to the agent as an influential precedent under the memory policy. When \(U_i\) is binary, \(U_i=1\) denotes a genuinely correct or useful episode and \(U_i=0\) denotes an incorrect, misleading, or harmful episode. The same notation also applies to continuous utility scores.

\subsection{The Echo Gap}
\label{subsec:echo-gap}

The Echo Gap is frequency-correlated reward inflation in a memory loop. It has two components.

First, the self-grade inflates wrong or low-utility memories:
\begin{equation}\label{eq:inflate}
\mathbb{E}[b_i \mid U_i = 0] > 0.
\end{equation}
The condition in Equation~\eqref{eq:inflate} says that incorrect episodes are scored as more useful than their true downstream value justifies.

Second, the inflation is operationally amplified by reuse among wrong memories, a positive covariance stated in Equation~\eqref{eq:reuse-cov}:
\begin{equation}\label{eq:reuse-cov}
\operatorname{Cov}(b_i,n_i \mid U_i=0) > 0.
\end{equation}
Equivalently, the empirical sections report the conditional association in Equation~\eqref{eq:reuse-corr},
\begin{equation}\label{eq:reuse-corr}
\operatorname{Corr}(b_i,n_i \mid U_i=0),
\end{equation}
which asks whether the wrong memories that are most overvalued are also the wrong 
memories whose high trust scores most strongly influence future decisions.

\begin{figure}[t]
    \centering
    \includegraphics[width=0.88\linewidth]{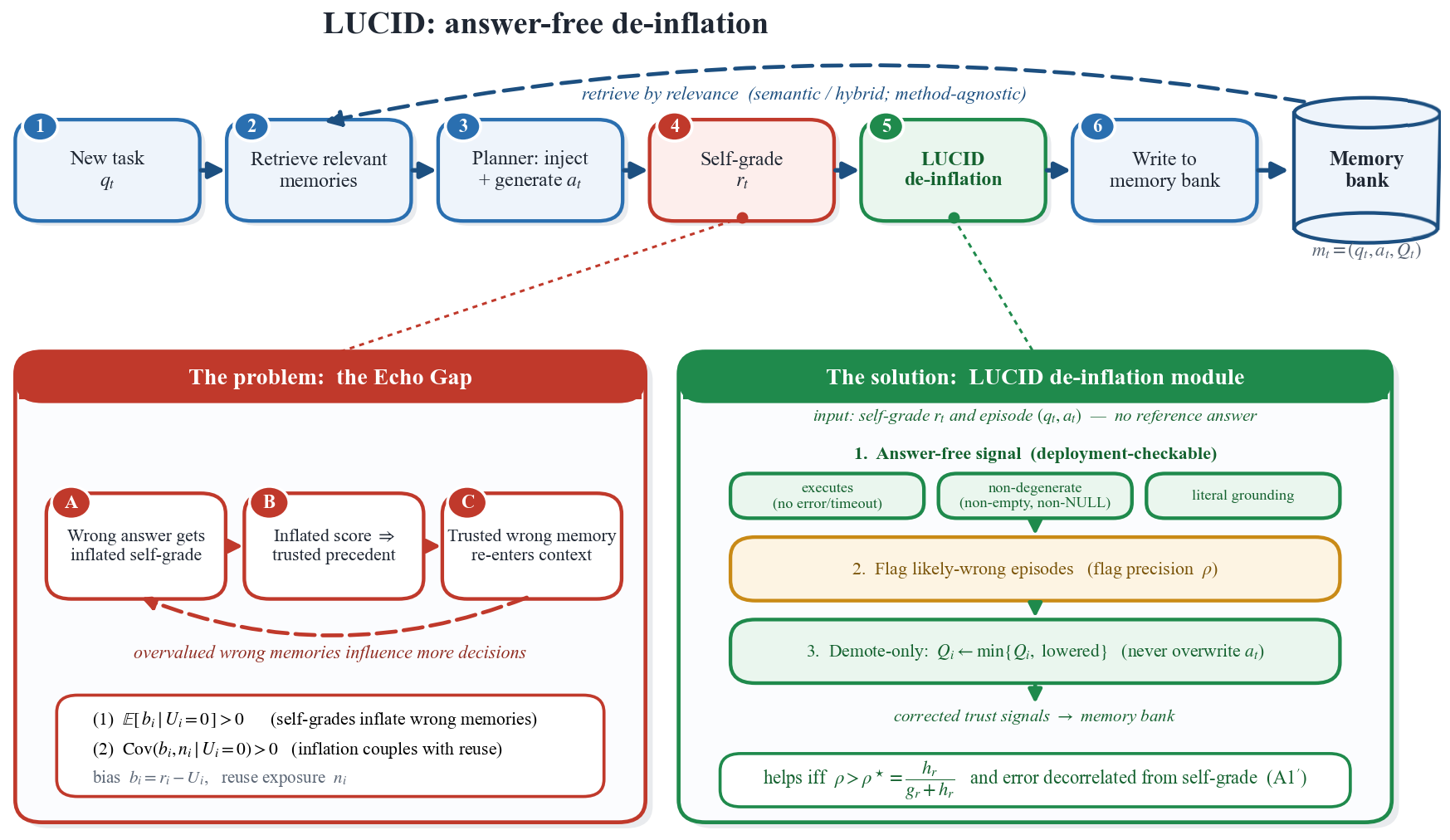}
    \caption{Conceptual mechanism of the Echo Gap in self improving memory agents and LUCID which demote the inflated past experiences}
    \label{fig:echo-gap-flow}
\end{figure}

Figure~\ref{fig:echo-gap-flow} summarizes this feedback mechanism.

Together, the two conditions in Equations~\eqref{eq:inflate} and~\eqref{eq:reuse-cov} distinguish the Echo Gap from ordinary miscalibration. A one-time calibration error can make reported scores numerically inaccurate without necessarily changing future behavior. The Echo Gap is a feedback loop: a wrong memory receives an inflated score, the memory exposes it to agent as correct solution, the agent conditions on it as a trusted precedent, and subsequent write-back can reinforce the same pattern. The failure is therefore not merely noisy memory, but inflation that is coupled to memory reuse.

\paragraph{Why the inflation compounds: a retrieval-and-trust amplification result.}
The covariance condition above is not merely a statistical curiosity; it provably amplifies the agent's exposure to wrong memories, and it does so through \emph{two} distinct channels. The influence a memory exerts on future behavior is the product of how often it is \emph{retrieved} and how strongly the agent \emph{trusts} it once retrieved. Write the influence of memory \(i\) as \(\mathrm{infl}(i)=\pi(i)\,\tau(Q_i)\), where \(\pi(i)\) is its retrieval probability and \(\tau(\cdot)\ge 0\) is a non-decreasing \emph{trust weight} with which the agent conditions on a retrieved memory of stored score \(Q_i\). In a Memento-style loop the stored self-grade is injected as exactly such a utility/trust annotation (Section~\ref{subsec:memory-loop-signals}). Let the \emph{reuse-weighted error mass} be \(M=\sum_{i:U_i=0}\mathrm{infl}(i)\), the share of influence spent on wrong memories. The two channels can both, separately, turn a static grading error into a compounding one.

\begin{theorem}[Echo-Gap amplification: retrieval and trust channels]
\label{thm:amplification}
Compare an honest bank (\(b_i{=}0\), so \(Q_i{=}U_i\)) with an inflated bank that raises the stored score of wrong memories to \(Q_i{=}b\in(0,1]\) while leaving correct memories at \(Q_i{=}1\); let \(W\) and \(R\) be the numbers of wrong and correct memories. The amplification \(A:=M_{\mathrm{infl}}/M_{\mathrm{hon}}\) factorizes as \(A=A_{\mathrm{ret}}\cdot A_{\mathrm{tr}}\) into a retrieval channel and a trust channel.
\begin{itemize}\itemsep1pt
\item[\emph{(i)}] \emph{(Retrieval channel.)} Under score-ranked softmax retrieval \(\pi(i)\propto e^{Q_i/T}\), \(T>0\) (top-\(k\) is the \(T\to0\) limit), with flat trust \(\tau\equiv\mathrm{const}\),
\begin{equation}\label{eq:amp}
1 \;\le\; A_{\mathrm{ret}} \;=\; \frac{e^{b/T}\,(W+Re^{1/T})}{We^{b/T}+Re^{1/T}} \;\le\; e^{b/T},
\end{equation}
with \(A_{\mathrm{ret}}\to e^{b/T}\) in the sparse-error limit \(W\ll Re^{1/T}\), and \(A_{\mathrm{ret}}=1\) under reward-blind (similarity-only) retrieval. The same upper bound \(A_{\mathrm{ret}}\le e^{b_{\max}/T}\) holds for heterogeneous inflation, with \(b_{\max}\) the largest per-memory inflation.
\item[\emph{(ii)}] \emph{(Trust channel.)} Under any retrieval rule whose probabilities \(\pi(i)\) do not depend on the stored score and in particular similarity-only retrieval, so that \(A_{\mathrm{ret}}=1\): if the agent conditions on retrieved memories through a trust weight that is strictly increasing on \(\{0,b\}\) with \(\tau(0)>0\), then
\begin{equation}\label{eq:amp-trust}
A_{\mathrm{tr}} \;=\; \frac{\tau(b)}{\tau(0)} \;>\; 1 .
\end{equation}
Inflation therefore amplifies the influence of wrong memories \emph{even when retrieval is reward-blind}: it does not change which memories are retrieved, but it raises the trust with which each retrieved wrong memory is reused, by the factor \(\tau(b)/\tau(0)\).
\end{itemize}
\end{theorem}

The proof is in Appendix~\ref{app:proofs}. Both bounds separate the Echo Gap from a one-shot grading error: the inflation does not average out, it is amplified \emph{multiplicatively} in how often and how trustingly wrong memories are reused. The two channels cover the two retrieval regimes the paper studies, so the theory is no longer confined to a regime the main experiments do not use. \emph{(i)~The retrieval channel} is the score-ranked memory-bank regime of Section~\ref{subsec:bank-level-evidence}: amplification reaches \(e^{b/T}\) and grows as retrieval becomes greedier (smaller \(T\)), and \(A_{\mathrm{ret}}=1\) only when retrieval ignores the stored score entirely. \emph{(ii)~The trust channel} is the regime the deployed agent actually occupies: the memory agent (BIRD experiment) loop retrieves by SimCSE similarity alone (\(A_{\mathrm{ret}}=1\)) yet the stored self-grade is shown to the agent as a trust/utility annotation and not used to re-rank (Section~\ref{subsec:bird-setup}), exactly the score-conditioned trust \(\tau(Q_i)\) the theorem assumes, so inflation still compounds through \(A_{\mathrm{tr}}=\tau(b)/\tau(0)>1\) rather than through retrieval mass. This resolves the apparent gap that under similarity retrieval ``\(A=1\)'': amplification under similarity retrieval is real, it simply runs through trust rather than ranking. The two channels are complementary and multiply (\(A=A_{\mathrm{ret}}A_{\mathrm{tr}}\)); a deployment can sit anywhere on the spectrum from pure-similarity (\(A_{\mathrm{ret}}{=}1\), trust-only) to pure-softmax (\(\tau\) flat, ranking-only).

The trust channel is not hypothetical: it is what the end-to-end agent \emph{measures}. In BIRD experiment the stored self-grade is binary and ``trusted'' means \(r{=}1\), so inflation moves a wrong memory from the untrusted set (\(\tau(0)\)) into the trusted set (\(\tau(1)\)); the memory-to-behavior coupling \(\kappa{\approx}0.38\) measured on the real agent (Section~\ref{subsec:bird-attractor}), error rising with the trusted-but-wrong fraction of the retrieved set, is precisely the empirical signature of \(A_{\mathrm{tr}}>1\), and the write-back dynamics of that trust amplification are analyzed in Theorem~\ref{thm:attractor}. The result thus formalizes, rather than replaces, the empirical coupling \(\operatorname{Cov}(b_i,n_i\mid U_i{=}0)>0\) reported below for the score-ranked banks and the coupling \(\kappa>0\) measured for the similarity-retrieval agent.

\subsection{What a de-inflation signal must satisfy: \texorpdfstring{\Aone{}}{EIA}}
\label{subsec:deinflation-criterion}
\label{subsec:deinflation-deploy}

De-inflation lowers the stored reward of memories whose grading appears inflated, so the agent stops trusting them as correct precedents. For this to help rather than hurt, the signal that drives the demotion must not reproduce the error pattern that created the inflated scores. Let \(V_i\) be a verifier score and \(\nu_i = V_i - U_i\) its error. A verifier satisfies \Aone{} on a validation bank if it both tracks truth and decorrelates its error from the self-grade bias, as in Equation~\eqref{eq:eia}:
\begin{equation}\label{eq:eia}
|\operatorname{Corr}(\nu, b)| < \tau_{\mathrm{dec}},
\qquad
\operatorname{Corr}(V, U) > \tau_{\mathrm{tru}}.
\end{equation}
The two requirements are independent: a verifier can be decorrelated from the bias yet uninformative about truth, or more truth-aligned on average while still erring in the same direction as the self-grade. The latter is the dangerous case, since demoting by such a signal can reintroduce the very distortion it was meant to remove. \Aone{} is an \emph{offline} signal-selection criterion: it reads \(U_i\) only on a validation bank to decide whether a candidate signal is trustworthy, after which the selected signal is used in label-free deployment. We treat it as directional rather than tied to a particular cutoff (the empirical separation is wide; Section~\ref{subsec:a1-make-or-break}). Appendix~\ref{app:eia-extended} discusses its status, a constructive sufficient condition under which decorrelation can be \emph{designed for} rather than hoped for, and why parametric re-graders fail to meet it.

\Aone{} is moreover a \emph{necessary} condition for de-inflation, not merely a description of a good verifier. Consider any de-inflation that pulls the stored score toward the verifier, \(Q_i'=Q_i-\alpha(Q_i-V_i)\), and decompose the verifier error as \(\nu=\beta b+\eta\) with \(\beta=\operatorname{Cov}(\nu,b)/\operatorname{Var}(b)\) and \(\operatorname{Cov}(\eta,b)=0\). Then (Proposition~2, proved in Appendix~\ref{app:proofs}) the corrected inflation has variance
\begin{equation}\label{eq:eia-necessity}
\operatorname{Var}(b')=\big(1-\alpha(1-\beta)\big)^2\operatorname{Var}(b)+\alpha^2\operatorname{Var}(\eta),
\end{equation}
so if the verifier echoes the self-grade strongly (\(\beta\ge1\)) then \(\operatorname{Var}(b')>\operatorname{Var}(b)\) for \emph{every} step \(\alpha\in(0,1]\) (demotion can only re-inject inflation), whereas the inflation removed at the optimal step,
\begin{equation}\label{eq:eia-payoff}
\operatorname{Var}(b)-\min_{\alpha}\operatorname{Var}(b')=\frac{(1-\beta)^2\operatorname{Var}(b)^2}{(1-\beta)^2\operatorname{Var}(b)+\operatorname{Var}(\eta)},
\end{equation}
is strictly increasing as the error decorrelates (\(\beta\to0\)) and as it vanishes (\(\operatorname{Var}(\eta)\to0\)). The recoverable payoff is thus a closed-form function of exactly \Aone{}'s two axes, so the criterion is \emph{derived} from the correction objective rather than posited.

\paragraph{Deployment and the precision condition.}
In deployment we apply de-inflation as a targeted de-inflation: a detector flags an episode as likely wrong using only answer-free evidence, and we lower that memory's stored reward, never re-solving the task nor using any ground-truth label. Because the intervention acts only on flagged episodes, a second quantity governs whether it helps: detector precision \(\rho=\Pr[U_i=0\mid\mathrm{flagged}]\). With expected per-flag gain \(g_r>0\) on truly wrong memories and loss \(h_r>0\) on good ones, the per-flag effect is \(\rho g_r-(1-\rho)h_r\), so de-inflation is beneficial in expectation iff
\begin{equation}\label{eq:rhostar}
\rho > \rho^\star := \frac{h_r}{g_r+h_r}.
\end{equation}
\Aone{} concerns the \emph{signal} (its error must be decorrelated, Equation~\eqref{eq:eia-payoff}); precision concerns is that de-inflated memories must be wrong often enough)=. An answer-free signal must be decorrelated by construction' its de-inflation precision we verify empirically. Both hold for the signal used in our experiments.

% 

% ============================================================================
% ============================================================================
\FloatBarrier
\subsection{Bank-Level Evidence}
\label{subsec:bank-level-evidence}

The Echo Gap studied to determine if it appears in live memory banks. The bank-level diagnose mechanism studied by asking whether label-free self-grading inflates wrong memories and whether, among wrong memories, the most overvalued ones receive more reuse exposure. Later sections study whether suitable verification signals can correct the bank and whether the same mechanism produces behavioral harm in a full agentic loop.

\paragraph{Factual memory banks.}
\label{subsec:factual-banks}

We begin with factual question-answering memories, where post-hoc correctness can be measured. Each memory contains a question, the agent's answer, and a self-grade written without access to the gold answer. Ground truth is used only after the memory has been written, to measure whether the self-grade matched correctness. This keeps the experiment in the label-free regime defined in Section~\ref{sec:echo-gap}.

The main bank uses Anthropic Claude Haiku 4.5 as the answering and self-grading agent. To test whether the phenomenon is tied to a single model family, we also construct self-graded banks from OpenAI GPT-5.4-mini and the frontier GPT-5.4 model on the same factual distribution. The purpose of this cross-vendor reproduction is not to produce an exact model leaderboard, because the gold-judging setup is not identical across all banks. Rather, it tests whether the same failure mode appears outside the original model family.

\paragraph{Metrics.}
\label{subsec:echo-gap-metrics}

Three quantities are reported. First, each memory's self-grade bias is computed in Equation~\eqref{eq:bias-bank}:
\begin{equation}\label{eq:bias-bank}
b_i = r_i - U_i .
\end{equation}
Positive bias means that the memory is valued more highly than its ground-truth utility justifies.

Second, for binary correctness, leniency is reported in Equation~\eqref{eq:leniency}:
\begin{equation}\label{eq:leniency}
\Pr[r_i=\mathrm{correct}\mid U_i=0],
\end{equation}
the probability that the model endorses its own wrong answer as correct. This is the headline severity measure: it asks how often wrong memories are admitted as if they were correct.

Third, the analysis measures whether the bias is operationally amplified by reuse among wrong memories, via the covariance in Equation~\eqref{eq:cov-bank}:
\begin{equation}\label{eq:cov-bank}
\operatorname{Cov}(b_i,n_i \mid U_i=0),
\end{equation}
and, for scale-free reporting, the corresponding conditional correlation in Equation~\eqref{eq:corr-bank}:
\begin{equation}\label{eq:corr-bank}
\operatorname{Corr}(b_i,n_i \mid U_i=0).
\end{equation}
Here \(n_i\) is the reuse count of memory \(m_i\) under the memory policy. The Echo Gap is present when wrong memories are overvalued and receive later reuse exposure.

\begin{table}[htbp]
\centering
\begin{tabular}{llcc}
\toprule
Model & Role & Leniency & 95\% CI \\
\midrule
Claude Haiku 4.5 (Anthropic) & self-grader & 0.31 & [0.26, 0.35] \\
GPT-5.4-mini (OpenAI) & self-grader & 0.54 & [0.49, 0.58] \\
frontier GPT-5.4 (OpenAI) & self-grader & 0.41 & [0.36, 0.47] \\
\bottomrule
\end{tabular}
\caption{Self-grading endorses wrong memories across model families. Leniency is \(\Pr[\mathrm{self}=\mathrm{correct}\mid U=0]\), pooled over three seeds. The table establishes reproduction of the failure mode, not an exact vendor comparison.}
\label{tab:leniency}
\end{table}

\paragraph{Self-grading inflates wrong memories.}
\label{subsec:self-grading-inflates}

Table~\ref{tab:leniency} reports leniency on the factual banks. The Claude Haiku 4.5 bank endorses 31\% of its own wrong facts, with a 95\% confidence interval of [26, 35]. The cross-vendor banks show the same qualitative pattern: GPT-5.4-mini endorses 54\% of its wrong answers, and the frontier GPT-5.4 model still endorses 41\%. These numbers should not be read as a direct model ranking, because the gold-judging setup is not identical across all banks. The robust conclusion is more conservative and more important: self-grade inflation appears in multiple model families, and it persists even at higher capability and LLM frontier models.

For these banks, the aggregate bias is positive, and the error is heterogeneous across topics. This matters because a uniform offset could in principle be handled by a simple calibration shift. The observed pattern is instead memory-specific: some wrong memories are much more inflated than others, while some high-scoring memories are genuinely useful. This heterogeneity is why later sections study per-memory correction rather than only global recalibration.

Under score-ranked reuse the inflation also couples to the memory policy: the bias and reuse count are positively associated among wrong memories, \(\operatorname{Cov}(b_i,n_i\mid U_i{=}0)>0\) (Equations~\eqref{eq:cov-bank}--\eqref{eq:corr-bank}), so the memories most promoted by the bank are disproportionately the ones whose utility was most overestimated. This is the bank-level signature of the Echo Gap, and it motivates the three questions examined next: whether an LLM-as-a-judge can re-grade and de-inflate the bank or merely reproduces the same bias (Section~\ref{sec:verifier-criteria}), why global calibration cannot (Section~\ref{sec:theory-repair}), and whether the same inflation degrades a full agentic loop (Section~\ref{sec:bird-end-to-end}).

% ============================================================================
\section{Verifier Criteria: When a Stronger Judge Is Not Enough}
\label{sec:verifier-criteria}
% ============================================================================

The natural response to bank-level inflation is to audit the bank with another model: a stronger LLM, a different-family LLM, or an LLM-as-a-judge re-grading the stored episodes. We test this against the \Aone{} criterion of Section~\ref{subsec:deinflation-criterion}, which a usable de-inflation signal must satisfy on both axes ($|\operatorname{Corr}(\nu,b)|$ small \emph{and} $\operatorname{Corr}(V,U)$ large), and then ask whether the same criterion predicts correction payoff once the bank is actually modified.

\subsection{Verifier candidates}
\label{subsec:verifier-candidates}

We evaluate six base verifiers on the same Haiku bank, spanning three classes, plus panels. (i)~\emph{Same-model, no-new-evidence} baselines: self-consistency by resampling and an adversarial same-model judge. (ii)~\emph{Cross-vendor parametric re-graders} from the GPT-5.4 family (nano, mini, largest called GPT-5.4 without any postfix), which test the intuitive stronger-judge remedy. (iii)~A \emph{retrieval-based information verifier} that conditions on external evidence rather than parametric re-evaluation, testing a different evidence channel. We also combine the cross-vendor judges into mean/majority/capability-weighted ensembles and a gold-fit optimal aggregator that upper-bounds any combination. All verifiers are scored post hoc; \(U_i\) is used only to compute the diagnostics \(\operatorname{Corr}(V,U)\) and \(\operatorname{Corr}(\nu,b)\), never given to the verifier.

\subsection{\texorpdfstring{\Aone{}}{EIA} Experiment}

\label{subsec:a1-make-or-break} 

Table~\ref{tab:a1} reports the result. The same-model baselines fail for different reasons. Self-consistency is nearly decorrelated from the original bias, but it does not move sufficiently toward truth. The adversarial same-model judge moves somewhat more toward truth, but its error remains correlated with the original self-grade bias. 

\begin{table}[htbp]
\centering\small
% \caption{\Aone{} make/break on the live Haiku bank, pooled over three seeds ($n{=}300$). PASS iff
% $|\Corr(\nu,b)|<0.3$ and $\Corr(V,U)>0.3$. Every \emph{parametric} re-grader fails, via two distinct
% routes; only the decorrelated information channel passes. frontier GPT-5.4 row from \cref{sec:ladder}.}
\begin{tabular}{lccl}
\toprule
Verifier & \(\operatorname{Corr}(\nu,b)\) & \(\operatorname{Corr}(V,U)\) & failure mode \\
\midrule
self-consistency (resample)        & $-0.03$ & $+0.10$ & weak truth tracking \\
adversarial same-model judge       & $+0.36$ & $+0.22$ & error correlated with self-grade \\
cross-vendor re-grade: GPT-5.4-nano   & $+0.11$ & $-0.04$ & not truth-tracking \\
cross-vendor re-grade: GPT-5.4-mini   & $+0.32$ & $+0.18$ & error correlated with self-grade \\
cross-vendor re-grade: frontier GPT-5.4 & $+0.31$ & $+0.29$ & weak on both \\
verifier ensemble (nano + mini + frontier, mean)   & $+0.33$ & $+0.20$ & shared bias survives averaging \\
\quad\emph{ensemble, optimal aggregator (upper bound)} & $+0.30$ & $+0.42$ & even the best panel of these judges fails \\
\textbf{retrieval-based information verifier}         & $+0.05$ & $+0.76$ & strongly truth-tracking, decorrelated \\
\bottomrule
\end{tabular}
\caption{\Aone{} on the live Haiku bank. Parametric re-graders fail via two routes: some are not truth-tracking, while stronger ones become more truth-aligned but remain error-correlated. The retrieval-based information verifier is the only tested signal that is strong on both axes: it tracks truth far more strongly than any parametric judge while keeping its error decorrelated.}
\label{tab:a1}
\end{table}

Cross-vendor parametric re-grading also fails, but the failure mode changes with capability. GPT-5.4-nano is roughly decorrelated but not truth-tracking. GPT-5.4-mini becomes more truth-aligned, but its error is correlated with the original bias. The frontier GPT-5.4 model(the biggest in size) is weak on both axes: it does not become strongly truth-tracking, and its error stays correlated with the original bias.

Combining the parametric judges into a panel did not help on this bank. A mean, majority, or capability-weighted vote inherits their shared bias: the two strongest judges' errors are themselves correlated ($\operatorname{Corr}(\nu_i,\nu_j)=+0.69$), so averaging reduces variance, not bias, leaving $\operatorname{Corr}(\nu,b)=+0.33$, indistinguishable from a single judge. Even the gold-fit optimal aggregator of all three judges (an upper bound that uses $U$ to combine them) reaches only $\operatorname{Corr}(V,U)=+0.42$ with $\operatorname{Corr}(\nu,b)=+0.30$, still short of the decorrelated channel on both axes; on this bank, unanimous agreement of the three judges is correct only $21\%$ of the time. Panels of judges that share the generator's blind spots therefore have few effective votes \citep{ninejudges2026}: on this bank, capability, count, and aggregation did not manufacture the decorrelation \Aone{} requires.

In contrast, the retrieval-based information verifier passes both conditions: its error is weakly correlated with the self-grade bias and its score strongly tracks truth. This is the only verifier in the tested set that satisfies \Aone{}.

\Aone{} is a \emph{property} whose passing class also contains the genuinely answer-free, imperfect BIRD signal (precision \(\rho{=}0.90\), Section~\ref{subsec:bird-answer-free-signal}), and Proposition~2 ties the recoverable payoff to a verifier's two correlations for \emph{any} signal, so the table establishes a \emph{contrast} (parametric re-graders never reach the decorrelated region), not a ceiling. The criterion is also genuinely \emph{predictive} rather than a post-hoc relabeling: under leave-one-seed-out cross-validation the retrieval channel is the unique \Aone{}-pass on the training seeds and the rank-one de-inflation signal on the held-out seed in all three folds (mean PASS-minus-FAIL gap \(+0.45\)). Both arguments are developed in Appendix~\ref{app:eia-extended}. The takeaway is that the useful intervention is not another score from a more capable model but another \emph{error channel}: increasing capability raises \(\operatorname{Corr}(V,U)\) without lowering \(|\operatorname{Corr}(\nu,b)|\), so a parametric judge sharing the generator's blind spots stays unsafe for correcting this bank. The next question is whether this diagnostic predicts actual correction payoff when the bank is modified.

% ============================================================================
\subsection{De-Inflation Payoff on the Bank}
\label{subsec:deinflation-payoff}

We next ask whether the \Aone{} criterion predicts what happens when we actually de-inflate the bank. Each verifier drives the \emph{same} conservative, de-inflation, no overwrite an answer, or re-solve a task (the full operator is given in Appendix~\ref{app:filtering}). Holding the operator fixed makes the comparison across verifiers turn entirely on signal quality. We score each corrected bank by the change in \(\operatorname{Corr}(Q,U)\), estimated by bootstrap; ground truth is used only for this evaluation, never by the verifier.

\paragraph{Correction payoff.}
\label{subsec:demotion-correction}
\label{subsec:correction-payoff}

Table~\ref{tab:payoff} reports the payoff for each verifier. The retrieval-based information verifier, the only non-oracle signal that satisfies \Aone{}, produces the largest non-oracle gain, lifting truth tracking from \(+0.22\) to \(+0.81\) and so recovering most of the oracle ceiling (\(+0.75\)). The parametric LLM re-graders do not provide reliable improvement: GPT-5.4-mini and the adversarial judge barely move the ranking (\(+0.03\)), the frontier GPT-5.4 grader yields only a small gain (\(+0.10\)), GPT-5.4-nano harms the bank (\(-0.16\)), and the judge ensemble behaves the same way. The label-blind global-calibration baseline, by construction, cannot change the ranking at all.

\begin{table}[htbp]
\centering
\small
\begin{tabular}{@{}p{0.42\linewidth}ccc@{}}
\hline
Audit verifier & before $\rightarrow$ after & gain & $\Pr[\Delta>0]$ \\
\midrule
retrieval-based information verifier & $+0.22 \rightarrow +0.81$ & $+0.59$ & $1.00$ \\
oracle gold & $+0.22 \rightarrow +0.97$ & $+0.75$ & $1.00$ \\
frontier GPT-5.4 re-grade & $+0.22 \rightarrow +0.32$ & $+0.10$ & $0.94$ \\
self-consistency & $+0.22 \rightarrow +0.28$ & $+0.06$ & $0.84$ \\
verifier ensemble (mean vote) & $+0.22 \rightarrow +0.27$ & $+0.05$ & $0.81$ \\
adversarial same-model judge & $+0.22 \rightarrow +0.25$ & $+0.03$ & $0.67$ \\
GPT-5.4-mini re-grade & $+0.22 \rightarrow +0.25$ & $+0.03$ & $0.68$ \\
verifier ensemble (majority vote) & $+0.22 \rightarrow +0.21$ & $-0.01$ & $0.46$ \\
GPT-5.4-nano re-grade & $+0.22 \rightarrow +0.06$ & $-0.16$ & $0.03$ \\
\midrule
global monotone calibration (label-blind) & $+0.22 \rightarrow +0.22$ & $+0.00$ & -- \\
\hline
\end{tabular}
\caption{De-inflation payoff on the live memory bank (pooled over three seeds, $n=300$). The reported quantity is the change in $\mathrm{Corr}(Q,U)$ after the demotion-only correction (Appendix~\ref{app:filtering}), with $\Pr[\Delta>0]$ from a $2000$-sample bootstrap}
\label{tab:payoff}
\end{table}

The payoffs track \Aone{} as predicted: the parametric re-graders, however capable, do not substitute for decorrelation. Importantly, the gain is not a generic memory-pruning effect: against no-generation controls on the same bank, confidence thresholding is a monotone no-op (Lemma~1) and \emph{budget-matched} random pruning at LUCID's exact budget \emph{harms} the bank ($-0.16$), whereas the decorrelated de-inflation results in gains $+0.59$. The effect is thus \emph{which} memories are de-inflated, not how many; Appendix~\ref{app:filtering} gives the full comparison.

So \Aone{} predicts what can recovers most of the oracle de-inflation payoff, and identifies the decorrelation property the agent's deployment signal must share. The next section explains why global score calibration cannot substitute for this per-memory evidence, and derives the precision condition under which answer-free de-inflation helps.

% ============================================================================
\section{Why Calibration Fails and When De-Inflation Helps}
\label{sec:theory-repair}
\label{sec:global-calibration}
% ============================================================================

This section develops the theory in three steps: an elementary lemma on why label-blind global calibration cannot repair heterogeneous, per-memory inflation (Section~\ref{subsec:global-calibration-main}); the paper's main theoretical result, a dynamical analysis showing that the write-back loop drives the bank to a \emph{corrupted attractor} whose parameters we later measure on BIRD benchmark (Section~\ref{subsec:attractor}). Extended derivations are in Appendix~\ref{app:proofs}.

\subsection{Global Calibration Cannot Selectively Correct the Bank}
\label{subsec:global-calibration-main}

The Echo Gap is heterogeneous: two memories can carry the same self-grade yet differ in correctness, so a label-blind map that depends only on \(Q_i\) must treat them identically.

\paragraph{Lemma 1.}
Let a label-blind global correction have the form \(\widetilde{Q}_i=g(Q_i)\). If there exist memories \(i\) and \(j\) with \(Q_i=Q_j\) but \(U_i\ne U_j\), then no such correction can assign them different corrected utilities, so global calibration cannot resolve per-memory inflation without memory-specific evidence; as a special case, a monotone \(g\) leaves the top-\(k\) retrieved set unchanged. The proof is in Appendix~\ref{app:proofs}.

Any correction that \emph{does} separate two equally-scored memories must therefore read per-memory evidence, at which point it is precisely the de-inflation signal we study. Empirically (Appendix~\ref{app:calibration}, Table~\ref{tab:calibration}), every label-blind monotone map leaves the ranking diagnostics exactly at their raw self-grade values, and even gold-fit maps (Platt, isotonic) a label-free system could not build improve them only marginally and stay far below the oracle.

\subsection{Inflation Drives a Corrupted Attractor}
\label{subsec:attractor}

Lemma~1 explains why calibration cannot repair the bank, but it is static and does not capture what makes the Echo Gap dangerous: the bank is not fixed, it is evolving over new tasks. We analyze a stylized mean-field model of the write-back loop and then measure its parameters on a real agent (Section~\ref{subsec:bird-attractor}); the value of the model is precisely that its parameters are measurable and its central prediction is falsifiable.

Let \(p_t\in[0,1]\) be the fraction of trusted (retrievable) memories that are wrong at round \(t\). The agent solves a new task by conditioning on retrieved memories, with error rate \(e(p)=\min(1,\,e_0+\kappa p)\), where \(e_0\) is its error rate on an uncorrupted bank (\(p{=}0\)) and \(\kappa\ge 0\) is the \emph{memory-to-behavior coupling}: how strongly retrieving corrupted memory degrades behavior. A wrong episode enters the trusted set with the grader's leniency \(\ell=\Pr[r{=}1\mid U{=}0]\), a correct one with its sensitivity \(s=\Pr[r{=}1\mid U{=}1]\), and the trusted bank relaxes toward the composition of its inflow, \(q(p)=\frac{e(p)\,\ell}{e(p)\,\ell+(1-e(p))\,s}\).

\begin{theorem}[Corrupted attractor]
\label{thm:attractor}
In the model above:
\begin{itemize}\itemsep1pt
\item[\emph{(i)}] \emph{(Compounding.)} The loop admits a fixed point \(p^\star\) solving \(\kappa(\ell-s)\,p^2+\big(e_0(\ell-s)+s-\kappa\ell\big)\,p-e_0\ell=0\); for \(\kappa>0\) it strictly exceeds the one-shot corruption \(p_0^\star=\frac{e_0\ell}{e_0\ell+(1-e_0)s}\) that inflation produces with no feedback.
\item[\emph{(ii)}] \emph{(Stability.)} The benign fixed point is locally stable iff \(q'(p^\star)=\dfrac{\kappa\,\ell\,s}{\big(e^\star(\ell-s)+s\big)^2}<1\); beyond a critical coupling \(\kappa^\star(\ell,s)\) it loses stability and the loop is driven toward the fully-corrupted state \(p^\star=1\).
\end{itemize}
\end{theorem}

The proof is in Appendix~\ref{app:proofs}. Part~(i) is the precise sense in which inflation \emph{compounds} rather than averaging out: feedback (\(\kappa>0\)) lifts the steady-state corruption above the one-shot value, so the bank is more corrupted than the grader's leniency alone would imply. Part~(ii) identifies a coupling threshold beyond which the benign regime ceases to exist. Section~\ref{subsec:bird-attractor} measures \(\kappa,\ell,s\) on BIRD, confirms that the predicted attractor matches the observed bank, and delimits which of these regimes the real agent occupies.

Both theorems are minimal mechanism models, chosen so their parameters are measurable and their central prediction is falsifiable; the empirical test is of the prediction, not of the softmax, mean-field, or linear-dose idealizations, and the conclusions (inflation compounds, demotion helps by lowering exposure) require only \(\kappa>0\) and monotonicity, both measured. Appendix~\ref{app:calibration} discusses these in depth and demonestrate that why relaxing it does not overturn the conclusions.

\subsection{Precision Governs Answer-Free De-Inflation}
\label{sec:answer-free-repair}

De-inflation has two requirements. First, its signal must be decorrelated (EIA required). Second, because de-inflation acts only when a detector flags an episode as likely harmful , the flagged set must be precise enough.

Let \(\rho=\Pr[U_i=0\mid \mathrm{flagged}]\) denote detector precision on the flagged set. Let \(g_r>0\) be the expected gain from de-inflating a truly harmful flagged memory, and let \(h_r>0\) be the expected loss from mistakenly de-inflating a useful memory. The expected effect is given by Equation~\eqref{eq:repair-effect2},
\begin{equation}\label{eq:repair-effect2}
\Delta_{\mathrm{repair}}=\rho g_r-(1-\rho)h_r.
\end{equation}

\paragraph{Proposition 1.}
Suppose de-inflation acts only on flagged episodes. If false flagging a truly wrong flagged memory yields expected gain \(g_r>0\), and mistakenly flagged  useful memory yields expected loss \(h_r>0\), then, by Equation~\eqref{eq:rhostar2}, de-inflation is beneficial in expectation whenever
\begin{equation}\label{eq:rhostar2}
\rho > \rho^\star := \frac{h_r}{g_r+h_r}.
\end{equation}

The threshold follows by requiring the expected de-inflation effect \(\rho g_r-(1-\rho)h_r\) to be positive; the derivation is given in Appendix~\ref{app:proofs}. The answer-free signal we deploy satisfies both conditions: it is decorrelated from the self-grade (\Aone{}) and, as we show, flags inflated memories with precision well above the threshold. The end-to-end BIRD experiment tests it directly.

% ============================================================================
\section{End-to-End Experiment}
\label{sec:bird-end-to-end}
% ============================================================================

This section tests whether the same mechanism has an observable behavioral consequence in a full agentic loop. the test is whether the self-grade's reward inflation measurably degrades a realistic memory agent, and whether an answer-free de-inflation of the self-grade can results in performance gain.

\subsection{Benchmark and agent loop}
\label{subsec:bird-setup}

We evaluate on the full BIRD text-to-SQL development set: $1{,}534$ questions over real SQLite databases. Performance is measured with BIRD's official execution accuracy (EX): the predicted SQL and reference SQL are executed, and their result sets are compared.

The agent implements a faithful Memento-style memory agent, with identical hyperparameters across each algorithm for a fair comparison. For each new question, it retrieves the top-\(k{=}4\) most similar past episodes with the same SimCSE encoder as the original Memento research paper (Section~\ref{subsec:bank-level-evidence}). It then conditions a planner on the retrieved episodes and produces a plan, and an executor turns that plan into a single SQL query. After producing the SQL, then the episode is graded without access to the reference answer, and the episode is written back into memory. Within each seed, all arms see the same task ordering. All roles (planner, executor, and self-grader) use the same model. Full reproducibility details (hyperparameters, sandboxing, caching, cost, and the released artifacts) are in Appendix~\ref{app:bird-repro}.

\subsection{Experiments}
\label{subsec:bird-arms}

We compare three algorithms under an identical task stream, retrieval depth, prompts, and scorer. The first is our Memento implementation: a faithful re-implementation of the Memento loop, left unchanged. It retrieves memories, conditions a planner on them, generates SQL, grades the result, and writes the episode back into memory for later use.

The second algorithm is LUCID, our answer-free de-inflation. It is the \emph{identical} Memento agent (same planner, executor, retriever, prompts, and task order) and differs only in the stored reward: LUCID \emph{de-inflate} the memories flagged by the answer-free signal (Section~\ref{subsec:bird-answer-free-signal}). LUCID also never overwrites a memory's content or re-solves the task using additional LLM generation.
The third algorithm is a no-memory control: the \emph{identical} architecture so the agent works from the given task alone.

\subsection{Answer-free de-inflation signal}
\label{subsec:bird-answer-free-signal}

The de-inflation detector reads only the candidate's own inputs and behavior (the question, the SQL it produced, and how that SQL executes) and never a reference query or a gold result. For text-to-SQL we instantiate it with three cheap channels: (i) \emph{execution}, when the query errors, times out, or returns non-deterministic results across identical runs; (ii) \emph{degeneracy}, when it executes but returns an empty or all-NULL result for a question that expects an answer; and (iii) \emph{literal grounding}, when it filters on an entity-like string literal that does not appear in the question, a direct fingerprint of a value copied from a different, wrongly trusted memory.
These three channels are not specific to SQL; they instantiate a general class of answer-free signals available in most agentic loops, where a stored episode can be checked against real world available signals without the gold answer (Appendix~\ref{app:terminology} lists illustrative examples in other domains). The precision and recall of this signal on the self-graded bank are reported in Appendix~\ref{app:bird-repro}.

\subsection{Main result}
\label{subsec:bird-main-result}

Table~\ref{tab:bird-end-to-end} reports the end-to-end results across two seeds (seed~0 and seed~1). Across both seeds no-memory agent reached (mean $52.4\%$), Memento algorithm reached (mean $54.0\%$). and LUCID algorithm reached a mean $56.9\%$. The no-memory agnet never queries memory, so it is order-invariant and identical across seeds. Per-seed 95\% confidence intervals for the paired LUCID$-$naive difference exclude zero ($+1.9$ [$+0.2,+3.6$] in seed~0; $+3.9$ [$+2.1,+5.6$] in seed~1.

Removing the memory inflation thus \emph{causally} raises accuracy, the de-inflation counterpart of an inflation-injection experiment, and the within-stratum dose-response of Section~\ref{subsec:bird-attractor}, where the coupling $\kappa$ stays positive inside every difficulty band, rules out the natural confound that harder questions both retrieve worse memory and are independently harder to solve.

\begin{table}[htbp]
\centering
\small
\begin{tabular}{@{}lcccc@{}}
\toprule
 & \multicolumn{3}{c}{Pass@1 (EX)} \\
\cmidrule(lr){2-4}
Arm & Seed 0 & Seed 1 & Mean \\
\midrule
no-memory control            & $52.4\%$ & $52.4\%$ & $52.4\%$ \\
naive self-graded memory     & $53.8\%$ & $54.2\%$ & $54.0\%$ \\
\textbf{LUCID} (answer-free de-inflation) & $\mathbf{55.7\%}$ & $\mathbf{58.1\%}$ & $\mathbf{56.9\%}$ \\
\bottomrule
\end{tabular}
\caption{End-to-end results on the full BIRD text-to-SQL development set (official execution accuracy, EX).}
\label{tab:bird-end-to-end}
\end{table}

\subsection{Validating the dynamical model on the agent}
\label{subsec:bird-attractor-main}

The BIRD traces let us measure the parameters of Theorem~\ref{thm:attractor} directly and test its central prediction. Agent error rises monotonically with the corruption of the retrieved set (slope $\kappa\approx0.38$, positive within every difficulty stratum), the direct empirical realization of the trust channel of Theorem~\ref{thm:amplification}(ii): retrieval here is similarity-only ($A_{\mathrm{ret}}{=}1$), so $\kappa>0$ isolates the planner trusting inflated memories more. Plugging the independently measured $(\kappa,\ell,s,e_0)=(0.38,0.76,0.90,0.32)$ into the theorem predicts a corrupted attractor $p^\star{=}0.45$, which matches the \emph{observed} trusted-bank corruption ($0.42$, within $0.03$), whereas a static no-loop account underpredicts it ($0.28$): the compounding is real and quantified ($1.6\times$ the one-shot value). The measured $\kappa$ places BIRD in the model's \emph{stable} regime. A generation-free retrieval-depth sweep confirms the inflation is a property of the grader, not of $k$. Full measurements, the depth sweep, and the dose-response figure are in Appendix~\ref{app:bird-loop}.

% ============================================================================
\section{Conclusion}
\label{sec:conclusion}
% ============================================================================

Self-improving LLM agents that learn from external memory are not only retrieval systems but reward-sensitive ones: stored scores determine how past episodes are reused, trusted, and reinforced. When those scores are self-graded, wrong answers can be assigned an inflated utility, and when that inflation couples to reuse, a feedback loop in which overvalued errors are repeatedly surfaced as useful precedents. We showed the Echo Gap on live factual banks across model families, proved that correcting it requires a signal that both tracks truth and avoids echoing the self-grade bias (the EIA), and demonstrated on BIRD text-to-SQL that an answer-free de-inflation algorithm LUCID, using only deployment-available evidence, consistently outperform other baselines.

The broader lesson is that label-free memory improvement requires evidence whose failure mode differs from the agent's own: a stronger self-evaluator may improve average judgments yet preserve the bias that distorted the bank. Safer self-improving agents should separate memory writing from memory trust, audit stored episodes with decorrelated, answer-free evidence, and de-inflate the memories that influence the agent behavior for future tasks.

% ============================================================================

\appendix
\section{Terminology: Label-Free, Gold-Free, and Answer-Free}
\label{app:terminology}

We use three related terms carefully.

A \emph{label-free memory loop} is one in which the ground-truth utility \(U_i\) is unavailable when the memory is written. This is the deployment regime: the agent must decide what to store and trust before the correct outcome is known.

A \emph{gold-free signal} is any signal computed without reading \(U_i\). It may still use deployment-available evidence such as retrieved documents, program execution, schema constraints, format checks, invariants, or consistency tests.

An \emph{answer-free de-inflation signal} is a stronger kind of gold-free signal: it detects likely wrongness without reading the reference answer or target output. Examples include execution failure, parse failure, type or schema violation, degenerate output, violated invariants, or metamorphic inconsistency under input-preserving rewrites.

Such signals are imperfect by design. They often have low recall, because many wrong outputs do not crash, violate a schema, or disagree with an alternative path in an obvious way. Low recall is not fatal for de-inflation. What matters is whether the flagged subset is precise enough to satisfy the threshold, and whether the signal's errors are decorrelated from the self-grade bias.

\begin{table}[htbp]
\centering\small
\begin{tabular}{@{}ll@{}}
\toprule
Agent domain & Example answer-free signals (no gold answer) \\
\midrule
research agent   & schema/format validity; citation resolves \& supports; cross-source agreement \\
text-to-SQL      & executes (no error/timeout); non-degenerate (non-empty, non-NULL); literal grounding \\
code agent       & compiles; runs; type-checks; property/metamorphic tests; no invariant violated \\
math / proof     & checker/lint; unit/dimension consistency; cross-path agreement \\
tool use / API   & well-formed output against the contract; idempotent on repeat \\
\bottomrule
\end{tabular}
\caption{The answer-free de-inflation signal is domain-general.}
\label{tab:answerfree-domains}
\end{table}

\section{Proofs and Derivations}
\label{app:proofs}

\paragraph{Proof of Theorem~\ref{thm:amplification}.}
\emph{(i)~Retrieval channel.} With \(\tau\) flat and \(\pi(i)\propto e^{Q_i/T}\), the honest bank places wrong memories at \(Q{=}0\) and correct memories at \(Q{=}1\), so the reuse-weighted error mass is \(M_{\mathrm{hon}}\propto\frac{W}{W+Re^{1/T}}\). The inflated bank raises wrong memories to \(Q{=}b\), giving \(M_{\mathrm{infl}}\propto\frac{We^{b/T}}{We^{b/T}+Re^{1/T}}\). Dividing yields the stated closed form for \(A_{\mathrm{ret}}\). Because \(b\ge 0\), the factor \(\frac{W+Re^{1/T}}{We^{b/T}+Re^{1/T}}\in(0,1]\), so \(A_{\mathrm{ret}}\le e^{b/T}\); and \(A_{\mathrm{ret}}\ge 1\) because raising wrong-memory scores can only increase their share of the (monotone, normalized) softmax mass. As \(W\ll Re^{1/T}\), both correction factors tend to \(1\) and \(A_{\mathrm{ret}}\to e^{b/T}\). Under reward-blind retrieval \(\pi(i)\) does not depend on \(Q_i\), hence not on \(b\), so the masses coincide and \(A_{\mathrm{ret}}=1\). For heterogeneous inflation, replacing each wrong memory's factor \(e^{b/T}\) by \(e^{b_i/T}\le e^{b_{\max}/T}\) preserves the same upper bound.

\emph{(ii)~Trust channel.} Now let \(\pi(i)\) be independent of the stored score (similarity-only retrieval), so the retrieval probabilities \(\pi(i)\) of any fixed wrong memory are identical in the honest and inflated banks. Influence is \(\mathrm{infl}(i)=\pi(i)\tau(Q_i)\), so the reuse-weighted error mass over wrong memories is \(M=\sum_{i:U_i=0}\pi(i)\tau(Q_i)\). The honest bank scores every wrong memory at \(Q{=}0\) and the inflated bank at \(Q{=}b\), hence
\[
A_{\mathrm{tr}}=\frac{M_{\mathrm{infl}}}{M_{\mathrm{hon}}}=\frac{\sum_{i:U_i=0}\pi(i)\,\tau(b)}{\sum_{i:U_i=0}\pi(i)\,\tau(0)}=\frac{\tau(b)}{\tau(0)}.
\]
Since \(\tau\) is non-decreasing and \(b\ge 0\), \(A_{\mathrm{tr}}\ge1\), with strict inequality whenever \(\tau\) is strictly increasing on \(\{0,b\}\) and \(b>0\). When retrieval probabilities and trust both depend on the score, the masses multiply and \(A=A_{\mathrm{ret}}A_{\mathrm{tr}}\). \hfill\(\square\)

\paragraph{Proof of Theorem~\ref{thm:attractor}.}
\emph{(i)} A fixed point satisfies \(p=q(p)=\frac{e\ell}{e\ell+(1-e)s}\) with \(e=e_0+\kappa p\). Writing \(D=e\ell+(1-e)s=e(\ell-s)+s\) and clearing the denominator, \(p\,D=e\ell\), i.e.\ \(p\big[(e_0+\kappa p)(\ell-s)+s\big]=(e_0+\kappa p)\ell\), which rearranges to \(\kappa(\ell-s)p^2+\big(e_0(\ell-s)+s-\kappa\ell\big)p-e_0\ell=0\). At \(\kappa=0\) this gives \(p_0^\star=\frac{e_0\ell}{e_0\ell+(1-e_0)s}\). Because \(e(p)\) is increasing in \(p\) and \(q\) is increasing in \(e\) (\(\partial q/\partial e=\ell s/D^2>0\)), the map \(q(\cdot)\) is increasing, so for \(\kappa>0\) the fixed point satisfies \(p^\star=q(p^\star)>p_0^\star\). \emph{(ii)} The iteration \(p_{t+1}=q(p_t)\) is locally stable at \(p^\star\) iff \(|q'(p^\star)|<1\). By the chain rule \(q'(p)=\tfrac{\partial q}{\partial e}\,e'(p)=\tfrac{\ell s}{D^2}\,\kappa\), giving the stated condition. As \(\kappa\) grows, \(q'(p^\star)\) increases past \(1\); beyond the resulting \(\kappa^\star(\ell,s)\) the interior fixed point is unstable and the iteration is driven to the boundary \(p=1\) (where \(e=1\) and \(q(1)=1\)).

\paragraph{Proof of Lemma 1.}
Because the correction depends only on \(Q_i\), equal stored scores imply equal corrected scores: if \(Q_i=Q_j\) then \(\widetilde{Q}_i=g(Q_i)=g(Q_j)=\widetilde{Q}_j\). If the two memories have different ground-truth utilities, any correction that treats them identically fails to separate them, so no label-blind global map can selectively demote the wrong memory while preserving the useful one; a successful repair must condition on information beyond the original stored score. For the special case of score-ranked retrieval: if \(g\) is strictly increasing then \(Q_i>Q_j \Leftrightarrow g(Q_i)>g(Q_j)\), so the induced order, and hence the top-\(k\) set, is unchanged; if \(g\) is only non-decreasing, a fixed tie-breaking rule independent of \(g\) resolves any collapsed ties consistently, again leaving the deterministic top-\(k\) set unchanged. Monotone global calibration therefore cannot remove a wrong memory from the top-\(k\) set unless some memory-specific signal changes the relative order.

\paragraph{Derivation of Proposition 1.}
The expected repair effect on the flagged set is given by Equation~\eqref{eq:prop1-delta},
\begin{equation}\label{eq:prop1-delta}
\Delta_{\mathrm{repair}}=\rho g_r-(1-\rho)h_r.
\end{equation}
Repair is beneficial when \(\Delta_{\mathrm{repair}}>0\), which by Equation~\eqref{eq:prop1-chain} gives
\begin{equation}\label{eq:prop1-chain}
\rho g_r-(1-\rho)h_r > 0
\quad\Longleftrightarrow\quad
\rho g_r > (1-\rho)h_r
\quad\Longleftrightarrow\quad
\rho(g_r+h_r)>h_r
\quad\Longleftrightarrow\quad
\rho>\frac{h_r}{g_r+h_r}.
\end{equation}
Thus, de-inflation can be beneficial even if the detector misses many harmful episodes, provided that the episodes it does flag are harmful with sufficiently high probability.

\paragraph{Proof of Proposition 2.}
Write \(\nu=\beta b+\eta\) with \(\beta=\operatorname{Cov}(\nu,b)/\operatorname{Var}(b)\) and \(\operatorname{Cov}(\eta,b)=0\). The corrected bias is \(b'=(1-\alpha)b+\alpha\nu=(1-\alpha)b+\alpha(\beta b+\eta)=\big(1-\alpha(1-\beta)\big)b+\alpha\eta\). Since \(\operatorname{Cov}(b,\eta)=0\),
\[
\operatorname{Var}(b')=\big(1-\alpha(1-\beta)\big)^2\operatorname{Var}(b)+\alpha^2\operatorname{Var}(\eta),
\]
which is Equation~\eqref{eq:eia-necessity}. \emph{(a)} If \(\beta\ge1\) then \(1-\beta\le0\), so for \(\alpha\in(0,1]\) the coefficient \(1-\alpha(1-\beta)=1+\alpha(\beta-1)\ge1\); hence the first term is \(\ge\operatorname{Var}(b)\) and the second is strictly positive (whenever \(\operatorname{Var}(\eta)>0\)), giving \(\operatorname{Var}(b')>\operatorname{Var}(b)\). \emph{(b)} Minimizing the quadratic \(\operatorname{Var}(b')\) over \(\alpha\) gives the stationary point \(\alpha^\star=\dfrac{(1-\beta)\operatorname{Var}(b)}{(1-\beta)^2\operatorname{Var}(b)+\operatorname{Var}(\eta)}\) (clamped to \([0,1]\)); substituting yields
\[
\operatorname{Var}(b)-\operatorname{Var}(b')\big|_{\alpha^\star}=\frac{(1-\beta)^2\operatorname{Var}(b)^2}{(1-\beta)^2\operatorname{Var}(b)+\operatorname{Var}(\eta)},
\]
which is Equation~\eqref{eq:eia-payoff}. It increases as \(\beta\to0\) and as \(\operatorname{Var}(\eta)\to0\), and at \(\beta{=}0,\ \operatorname{Var}(\eta){=}0\) (i.e.\ \(V{=}U\)) equals \(\operatorname{Var}(b)\), removing all inflation. \hfill\(\square\)

% [Removed here] An earlier draft repeated the global-calibration argument (the
% "global score map", "monotone calibration", "heterogeneous inflation", and
% "connection to EIA" subsections) in this appendix. That material duplicated
% Lemma 1 and its discussion in Section~\ref{subsec:global-calibration-main}
% and the formal proofs in Appendix~\ref{app:proofs}, so it has been removed.

\subsection{Answer-Free De-Inflation and the Precision Threshold}
\label{sec:answer-free-repair-extended}

The de-inflation objective is to reduce the effective mass of wrong-but-trusted memories and its flagged set must be precise. This subsection formalizes the precision condition.

\paragraph{Why precision, not just decorrelation.}
\label{subsec:why-precision}

\Aone{} ensures the de-inflation signal does not merely echo the inflation it was meant to remove. But because de-inflation acts only on a flagged subset, demoting the stored reward of a flagged memory (from $1$ to $0$) and never overwriting its content or re-solving the task, a second quantity matters: the precision of the flagged set. A detector that is decorrelated on average can still hurt if the specific episodes it flags are mostly fine.

\paragraph{A precision condition for beneficial de-inflation.}
\label{subsec:repair-precision-threshold}

Let \(\rho\) denote the precision of the detector on the flagged set, as in Equation~\eqref{eq:precision-def-ext}:
\begin{equation}\label{eq:precision-def-ext}
\rho = \Pr[U_i=0 \mid \mathrm{flagged}].
\end{equation}
Here, \(U_i=0\) means that the episode is genuinely harmful or wrong for the purpose of reuse. Let \(g_r>0\) be the expected gain from repairing a truly harmful flagged episode, and let \(h_r>0\) be the expected loss from mistakenly repairing a useful episode. Then the expected effect of repair on a flagged episode is given by Equation~\eqref{eq:repair-effect-ext},
\begin{equation}\label{eq:repair-effect-ext}
\Delta_{\mathrm{repair}}
=
\rho g_r - (1-\rho)h_r.
\end{equation}
Repair is beneficial when \(\Delta_{\mathrm{repair}}>0\), which gives the threshold in Equation~\eqref{eq:rhostar-ext},
\begin{equation}\label{eq:rhostar-ext}
\rho > \rho^\star := \frac{h_r}{g_r+h_r}.
\end{equation}

This is Proposition~1 (Section~\ref{sec:answer-free-repair}), derived in Appendix~\ref{app:proofs}.

The condition is intentionally precision-based rather than recall-based. A low-recall detector may leave many bad memories untouched, but it can still improve the agent if the subset it does flag is sufficiently enriched for harmful episodes. This is why precision, not recall, governs de-inflation.

\paragraph{Answer-free signals.}
\label{subsec:answer-free-signals-repair}

The detector used for de-inflation need not observe the reference answer. These signals are imperfect. They may have low recall, because many wrong outputs do not trigger any obvious failure. However, low recall is not fatal for de-inflation. The precision threshold shows that a detector can still help if the subset it flags is sufficiently enriched for harmful episodes.

\paragraph{\texorpdfstring{\Aone{} and precision are complementary.}{A1 and precision are complementary.}}
\label{subsec:no-contradiction-a1}

The two conditions act at different levels. \Aone{} is a property of the \emph{signal}: its errors must be decorrelated from the self-grade bias, or demotion echoes the inflation it was meant to remove. Precision is a property of the \emph{flagged set}: the episodes actually demoted must be wrong often enough to clear the threshold.

\section{Extended Discussion of \texorpdfstring{\Aone{}}{EIA}}
\label{app:eia-extended}

This appendix collects material supporting the \Aone{} criterion of Section~\ref{subsec:deinflation-criterion} and the verifier experiment of Section~\ref{subsec:a1-make-or-break}.

\paragraph{A constructive sufficient condition for \texorpdfstring{\Aone{}}{EIA}.}
\Aone{} is not only checkable offline; it admits a \emph{constructive sufficient condition} that shows decorrelation can be designed for rather than merely hoped for. The self-grade \(r_i\) is a function of the generator's internal computation \(Z_i\) (its parametric beliefs about the task), so its error \(b_i=r_i-U_i\) is \(Z_i\)-measurable. If the verifier reads a channel \(W_i\) that is conditionally independent of that computation given the ground truth, \(W_i \perp Z_i \mid U_i\), then \(\nu_i=V_i-U_i\) is \(W_i\)-measurable and \(\operatorname{Cov}(\nu_i,b_i\mid U_i)=0\), so \(|\operatorname{Corr}(\nu,b)|\) is bounded by the truth-induced component alone. The retrieval channel attains \(\operatorname{Corr}(\nu,b)=+0.05\) (Table~\ref{tab:a1}). Its value is explanatory: it identifies \emph{why} a different information channel tends to satisfy \Aone{}, whereas parametric re-graders (stronger models, different families, or ensembles) reuse \(Z\)-correlated computation and do not.

\paragraph{On the status of \texorpdfstring{\Aone{}}{EIA}.}
the choice of thresholds $\tau_{\mathrm{dec}},\tau_{\mathrm{tru}}$ is not load-bearing. Proposition~2 makes the \emph{property} \Aone{} names, low error correlation and high truth tracking, a necessary condition for de-inflation to help; what is directional is only the \emph{binary cutoff} drawn on that property, not the property itself. The pass/fail \emph{threshold} is a diagnostic convenience whose role is to summarize which signal recovers the de-inflation payoff, and the empirical separation is wide rather than marginal. The channel we use sits at $|\Corr(\nu,b)|{=}0.05$, $\Corr(V,U){=}0.76$, whereas every parametric signal that tracks truth at all ($\Corr(V,U){>}0$) has $|\Corr(\nu,b)|\ge0.30$ and $\Corr(V,U)\le0.42$ (Table~\ref{tab:a1}). Any cutoff inside that gap induces the \emph{same} pass/fail partition, so the conclusion does not hinge on a particular $\tau$. The actual go/no-go quantity is not $\tau$ but the precision threshold $\rho^\star{=}h_r/(g_r{+}h_r)$ (Section~\ref{subsec:deinflation-deploy}), which is \emph{derived} from the de-inflation gain/loss trade-off rather than chosen. The constructive condition $W\perp Z\mid U$ is an \emph{explanatory design heuristic} (it explains why reading a different information channel \emph{tends} to decorrelate the error), not a guarantee assumed here. the paper never claims it holds exactly but measures the residual error correlation ($\Corr(\nu,b){=}0.05$), and because de-inflation acts only through the flagged set, Proposition~1 shows that a mildly correlated signal still helps as long as its flags clear $\rho^\star$.

\paragraph{The criterion does not rest on a near-oracle verifier.}
\Aone{} is a \emph{property}, and its passing class is not limited to near-oracles: the genuinely answer-free signal deployed end-to-end on BIRD (Section~\ref{subsec:bird-answer-free-signal}) is decorrelated from the self-grade \emph{by construction} (it reads program behavior, not parameters), yet is realistically imperfect, flagging wrong memories at precision \(\rho{=}0.90\) rather than near-\(1.0\). That signal is the one the deployment claim rests on. Also Proposition~2 shows \Aone{} governs the recoverable payoff for \emph{any} verifier through its two correlations, so the criterion is meaningful independently of whether any single tested verifier happens to be strong.

\paragraph{\Aone{} is a predictive selection criterion, not a restatement.}
Because \Aone{} (``a verifier helps when it tracks truth and decorrelates from the bias'') is close to a definition of a good verifier, it could in principle merely re-describe, after the fact, which signal happened to work on this bank. We test it as a genuine out-of-sample predictor by leave-one-seed-out cross-validation over the three banks ($n{=}100$ each): \Aone{} is evaluated only on two \emph{training} seeds to make a pass/fail selection, and the de-inflation payoff (Section~\ref{subsec:deinflation-payoff}) is then measured on the \emph{held-out} seed. The selection is stable and predictive: the retrieval channel is the unique \Aone{}-pass on the training seeds in all three folds, and on the held-out seed it is the rank-one de-inflation signal in all three folds, with a mean PASS-minus-FAIL payoff gap of $+0.45$. The partition is threshold-independent: the passing channel's \Aone{} margin ($\operatorname{Corr}(V,U)-|\operatorname{Corr}(\nu,b)|=+0.71$) is an order of magnitude above the next signal's ($+0.06$), so any cutoff inside that gap induces the same split. We do \emph{not} claim that \Aone{} finely ranks the \emph{failing} signals out-of-sample; their payoffs are statistically indistinguishable from zero, so their relative order is noise, and the operative output of \Aone{} is a binary selection of which signal to deploy, not a fine ranking.

\section{Generic Memory Filtering}
\label{app:filtering}

\paragraph{LUCID versus generic memory filtering.}
The decisive test of whether LUCID is more than a memory-management heuristic is how it compares to the memory-pruning family on the same bank (Table~\ref{tab:filtering}; pooled three-seed bank, $144$ trusted ($r{=}1$) memories of which only $15$ are genuinely correct). Confidence (self-grade) thresholding is a monotone no-op (Lemma~1). Self-consistency (uncertainty) pruning demotes the $81$ trusted memories it flags as uncertain and recovers only $+0.06$ (far below LUCID) while still demoting correct memories, consistent with its weak truth-tracking in Table~\ref{tab:a1}. The sharpest control is \emph{budget-matched} random pruning: at LUCID's exact budget ($K{=}123$), random pruning \emph{harms} the bank ($-0.16$) because it demotes $13$ of the $15$ correct memories, whereas LUCID demotes the same number of memories with zero collateral and gains $+0.59$. The gain is therefore attributable to \emph{which} memories are demoted, the decorrelated selection, not to how many. LUCID dominates random pruning at every budget tested (at $25/50/75/100\%$ of $K$: $+0.06/+0.13/+0.26/+0.59$ versus $-0.04/-0.08/-0.11/-0.16$), so it is not an artifact of the budget choice. Two cost-matched baselines that add \emph{no} LLM generation also fail to de-inflate: clipping the retrieval similarity floor leaves $\mathrm{Corr}(Q,U)\le+0.20$ across floors (it acts on relevance, not reward), and confidence thresholding is the monotone no-op above; every \emph{effective} alternative instead adds per-memory generation (a judge, a panel, or resampling) yet fails \Aone{} on this bank.
The small base of correct memories here ($15$) also makes the ``$0$ demoted'' figure a low-count estimate, and we read it accordingly.

\begin{table}[htbp]
\centering\small
\begin{tabular}{@{}lccc@{}}
\toprule
Memory-filtering rule & demoted & $\Delta\,\mathrm{Corr}(Q,U)$ & correct demoted (of 15) \\
\midrule
confidence / self-grade threshold          & $0$   & $+0.00$ & $0$ (monotone no-op) \\
uncertainty prune (self-consistency)       & $81$  & $+0.06$ & $4$ \\
random prune (matched budget)              & $123$ & $-0.16$ & ${\approx}\,13$ \\
\textbf{LUCID} (answer-free, decorrelated) & $123$ & $\mathbf{+0.59}$ & $\mathbf{0}$ \\
\bottomrule
\end{tabular}
\caption{LUCID versus generic memory filtering on the pooled three-seed bank ($n{=}300$); $144$ trusted memories, of which only $15$ are genuinely correct.}
\label{tab:filtering}
\end{table}

\section{Global Calibration Evidence and Modeling Assumptions}
\label{app:calibration}

\paragraph{Global calibration cannot repair the bank.}
Table~\ref{tab:calibration} confirms Lemma~1 on a focused SimpleQA factual bank ($N{=}50$), reporting three rank-based ranking diagnostics: Spearman $\rho(Q,U)$, AUC, and the top-10 gold rate. Because a monotone map preserves the retrieval order, every label-blind map (mean-subtraction, z-scoring, min--max rescaling, temperature scaling, sigmoid squash) leaves all three diagnostics \emph{exactly} at their raw self-grade values ($\rho{=}-0.08$, $\mathrm{AUC}{=}0.41$, top-10 gold ${=}0.00$), independent of the specific parameter settings, which are therefore not tuned. Even \emph{gold-fit} maps (Platt logistic, isotonic regression), which a label-free deployment could not construct because they require the unavailable labels, improve the ranking only marginally ($\rho$ to $+0.08$ and $+0.10$ respectively, against $+1.00$ for the oracle) and isotonic still retrieves zero gold memories in its top-10, exactly as Lemma~1 predicts: a score-only map cannot separate equal-scored memories of differing correctness. These rank-based metrics are on a separate, smaller bank and are not directly comparable to the Pearson $\mathrm{Corr}(Q,U)$ payoff on the larger bank in Table~\ref{tab:payoff}.

\begin{table}[htbp]
\centering\small
\begin{tabular}{@{}lccc@{}}
\toprule
Score map & Spearman $\rho(Q,U)$ & AUC & top-10 gold \\
\midrule
raw self-grade (identity)                 & $-0.08$ & $0.41$ & $0.00$ \\
\quad + mean-subtract / z-score / min--max & $-0.08$ & $0.41$ & $0.00$ \\
\quad + temperature ($T{=}2$) / sigmoid    & $-0.08$ & $0.41$ & $0.00$ \\
\midrule
Platt logistic \emph{(gold-fit)}          & $+0.08$ & $0.59$ & $0.10$ \\
isotonic \emph{(gold-fit)}                & $+0.10$ & $0.60$ & $0.00$ \\
\midrule
oracle (true utility)                     & $+1.00$ & $1.00$ & $0.40$ \\
\bottomrule
\end{tabular}
\caption{Ranking diagnostics under global score calibration. Rows group the identity (raw self-grade) and label-blind monotone maps (which share one row of values), gold-fit maps (Platt, isotonic), and the oracle. Columns: Spearman $\rho(Q,U)$, AUC, and the top-10 gold rate.}
\label{tab:calibration}
\end{table}

\section{BIRD Reproducibility, Hyperparameters, and Detector Precision}
\label{app:bird-repro}

\paragraph{Models and availability.}
All models used in this paper are publicly available production models, and we pin fixed versions so the experiments reproduce against the same snapshots. The answering and self-grading agent for the factual banks is Anthropic Claude~Haiku~4.5 (claude-haiku-4-5-20251001). The cross-vendor re-graders are the OpenAI GPT-5.4 family: gpt-5.4-nano-2026-03-17, gpt-5.4-mini-2026-03-17, and the frontier gpt-5.4-2026-03-05. Retrieval uses the rinceton-nlp/sup-simcse-bert-base-uncased SimCSE sentence encoder, the same retriever as Memento~\citep{memento2025,simcse2021}.

\paragraph{Reproducibility and hyperparameters.}
The three baseline algorithms share a single configuration; only the stored reward differs. Retrieval ranks the top-$k{=}4$ neighbors by SimCSE cosine similarity \emph{alone} (Memento-faithful, so stored labels are shown to the planner but never used to re-rank) and tags cases drawn from the same database. The case memory is unbounded and append-only: every episode is written back, so the bank grows across the run to the full task stream (up to $1{,}534$ episodes). Each candidate query runs in an isolated sandbox with a $30$-second timeout and is executed twice to detect nondeterminism. The reference query is executed only to compute official EX and is never visible to the agent. Two seeds ($0$ and $1$) are reported, and each seed fixes one task permutation shared by all algorithms. The end-to-end study spans three algorithms $\times$ two seeds $\times$ $1{,}534$ episodes, and each episode issues three model calls (planner, executor, self-grader). The answer-free de-inflation signal makes \emph{no} model call, so LUCID and the Memento agent baseline incur identical per-episode generation cost. This is on the order of $2.8{\times}10^{4}$ generation calls for the full evaluation, and fewer in practice because content-addressed caching replays the shared prefix at zero cost. SimCSE embedding is cheap relative to generation, and sandboxed SQL execution, not model inference, dominates wall-clock. All code, per-episode memory traces, exact run configurations, and result files are released at \url{https://github.com/MohammadAsadolahi/Reliable-Memory-Agents-in-the-Wild}.

\paragraph{Detector precision and recall.}
Measured directly on the self-graded bank, this signal flags wrong memories with precision $\rho=0.90$ (pooled, CI $[0.875,0.920]$, $n{=}649$ flags) against a base rate of $\approx0.46$, clearing the break-even threshold $\rho^\star{=}0.5$ (Proposition~1) by a wide margin while reading no reference answer; recall is low but harmless for de-inflation. The per-seed precision and recall figures follow. The answer-free signal flags wrong memories with precision $\rho=0.93$ in each of the two seeds run in the deployed configuration (95\% CIs $[0.89,0.96]$ and $[0.87,0.98]$; pooled across all three available banks $\rho=0.90$, CI $[0.875,0.920]$, $n{=}649$ flags), against a base rate of wrong memories of $\approx0.46$.

\section{Measuring the Loop on the Real Agent}
\label{app:bird-loop}
\label{subsec:bird-attractor}

Theorem~\ref{thm:attractor} predicts a corrupted attractor, but only if its parameters are real. The BIRD traces allow them to be estimated directly and the model's central prediction to be tested against the observed bank.

\begin{figure}[t]
\centering
\includegraphics[width=0.82\linewidth]{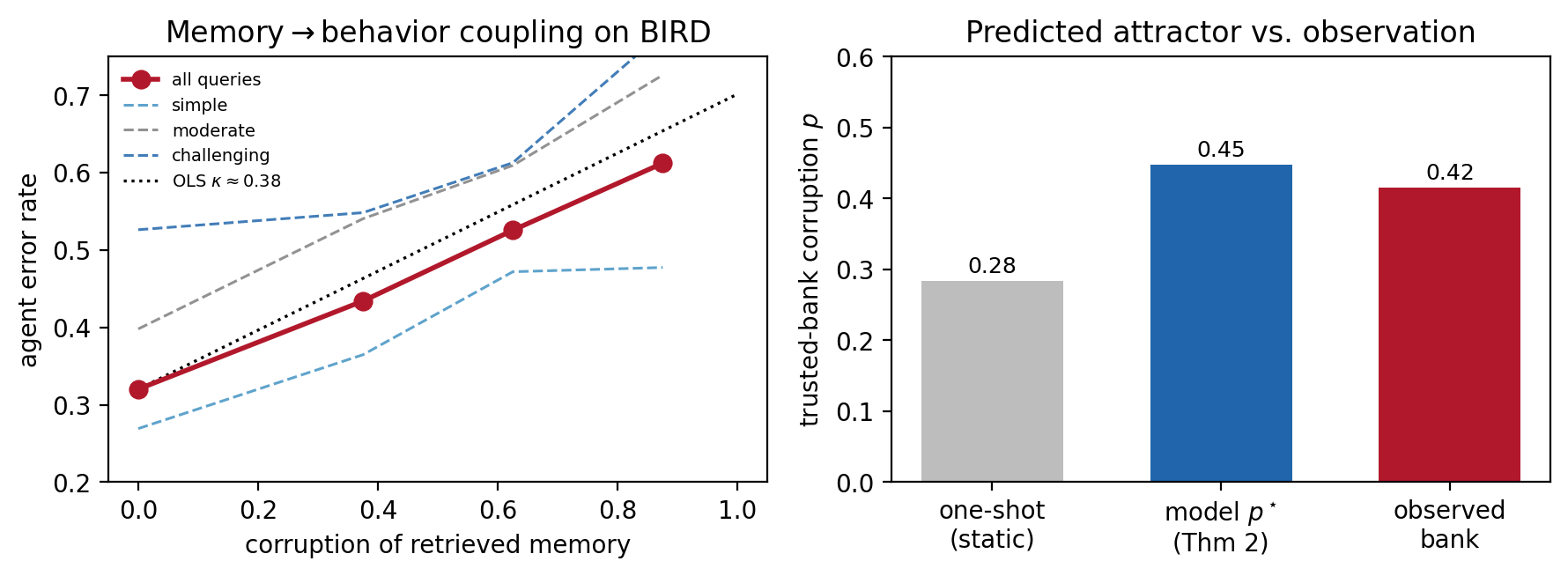}
\caption{Measuring the memory-to-behavior coupling on BIRD. Agent error rate rises monotonically with the corruption of the retrieved memory set (left), with slope $\kappa\approx0.38$ that persists within every difficulty stratum, so the coupling is not a difficulty artifact. Plugging the measured $(\kappa,\ell,s,e_0)$ into Theorem~\ref{thm:attractor} predicts a corrupted attractor $p^\star{=}0.45$ that matches the observed trusted-bank corruption $0.42$, while the static (no-loop) account underpredicts it at $0.28$ (right).}
\label{fig:kappa}
\end{figure}

\paragraph{The memory-to-behavior coupling is measurable and positive.}
For each query we compute the corruption of its retrieved set (the fraction of retrieved memories that are trusted yet, post hoc, actually wrong) and relate it to whether the agent then errs. Agent error rises monotonically with retrieved corruption, from $0.32$ when the retrieved set is clean to $0.61$ when it is fully corrupted (Figure~\ref{fig:kappa}), an ordinary-least-squares slope of $\kappa\approx0.38$. The slope stays positive \emph{within} every difficulty stratum ($\kappa=0.32,0.40,0.34$ for simple, moderate, and challenging questions), so it is not an artifact of harder questions both retrieving worse memory and being harder to solve. From the same run, the self-grader has leniency $\ell=0.76$ and sensitivity $s=0.90$, and the uncorrupted-bank error is $e_0=0.32$. This positive coupling is the direct empirical realization of the trust channel of Theorem~\ref{thm:amplification}(ii): because retrieval here is similarity-only ($A_{\mathrm{ret}}{=}1$), the harm of wrong memories cannot come from score-ranked retrieval mass, so $\kappa>0$ isolates the effect of the planner trusting inflated memories more, i.e.\ $A_{\mathrm{tr}}=\tau(1)/\tau(0)>1$ realized as a measured slope rather than assumed.

\paragraph{The predicted attractor matches the observed bank.}
Substituting these \emph{independently measured} values into Theorem~\ref{thm:attractor} yields a closed-loop corrupted attractor $p^\star=0.45$. The \emph{observed} corruption of the trusted bank, the fraction of self-graded-correct memories that are in fact wrong, is $0.42$, within $0.03$ of the prediction, and the corruption trajectory rises over the task stream before settling near this value. A static account that ignores the loop underpredicts the corruption at $0.28$; it is the dynamical term that reconciles prediction with observation, and the compounding it describes is real and quantified (the attractor is $1.6\times$ the one-shot value). 

\paragraph{Scope of the dynamical claim.}
The measured $\kappa=0.38$ places BIRD firmly in the model's \emph{stable} regime: corruption is bounded but materially elevated, and the loop does not collapse. Theorem~\ref{thm:attractor}(ii) predicts that agents with stronger memory dependence (larger $\kappa$) would cross the stability threshold. LUCID acts on the complementary lever, by demoting flagged memories it lowers the corruption the planner is exposed to, shifting retrievals toward the clean end of the dose-response curve in Figure~\ref{fig:kappa}.

\paragraph{Sensitivity to retrieval depth.}
Because re-running generation at every depth would require new model calls, we probe $k$-sensitivity \emph{generation-free}: re-embedding the bank reproduces the agent's stored retrieval exactly ($1{,}533/1{,}533$ top-$4$ sets matched), so at each depth we recompute the corruption of the retrieved set the planner conditions on and its coupling to behavior. Across $k\in\{1,2,4,8\}$ the retrieved-set corruption is essentially flat ($0.36,0.34,0.34,0.34$), confirming the inflation is a property of the grader, not of the depth, while the memory-to-behavior coupling \emph{strengthens} monotonically ($\kappa=0.15,0.22,0.38,0.44$) and the clean-bank error falls ($e_0=0.41,0.39,0.33,0.31$). Deeper retrieval therefore helps more when the bank is clean and costs more when it is corrupted. The error gap between fully-corrupted and clean retrieval widens from $+0.15$ at $k{=}1$ to $+0.33$ at $k{=}8$, all within the model's stable regime. The deployed $k{=}4$ is representative rather than a favorable choice (its slope is the headline $\kappa{=}0.38$), and the value of de-inflation, if anything, grows with retrieval depth.

\bibliographystyle{unsrtnat}
\bibliography{jobname}

\end{document}